\documentclass{article}

\usepackage[final,nonatbib]{neurips_2020}

\usepackage[utf8]{inputenc} % allow utf-8 input
\usepackage[T1]{fontenc}    % use 8-bit T1 fonts
\usepackage{hyperref}       % hyperlinks
\usepackage{bm}             % boldface math
\usepackage{url}            % simple URL typesetting
\usepackage{booktabs}       % professional-quality tables
\usepackage{amsfonts}       % blackboard math symbols
\usepackage{nicefrac}       % compact symbols for 1/2, etc.
\usepackage{microtype}      % microtypography

\usepackage{authblk}        % author list formatting

\usepackage{graphicx}
\usepackage{wrapfig,lipsum,booktabs}
\usepackage{multirow}
\usepackage{subfig}
\usepackage{enumitem}
\setlist[itemize]{leftmargin=*}

\makeatletter
\newcommand{\printfnsymbol}[1]{%
  \textsuperscript{\@fnsymbol{#1}}%
}
\makeatother

\makeatletter
\def\@fnsymbol#1{\ensuremath{\ifcase#1\or \dagger\or \ddagger\or
\mathsection\or \mathparagraph\or \|\or **\or \dagger\dagger
\or \ddagger\ddagger \else\@ctrerr\fi}}
\makeatother

\title{Interactive Teaching for Conversational AI}

\author{Qing Ping\thanks{These authors contributed equally to this work} }
\author{Feiyang Niu\printfnsymbol{1}}
\author{Govind Thattai\printfnsymbol{1}}
\author{Joel Chengottusseriyil}
\author{Qiaozi Gao}
\author{Aishwarya Reganti}
\author{Prashanth Rajagopal}
\author{Gokhan Tur}
\author{Dilek Hakkani-Tur}
\author{Prem Natarajan}

\affil{Alexa AI, Amazon \authorcr
  \{\tt pingqing, nfeiyan, thattg, jchengot, qzgao, areganti, rajagpra, gokhatur, hakkanit, premknat\}@amazon.com}

\begin{document}

\maketitle

\begin{abstract}
Current conversational AI systems aim to understand a set of pre-designed requests and execute related actions, which limits them to evolve naturally and adapt based on human interactions. Motivated by how children learn their first language interacting with adults, this paper describes a new Teachable AI system that is capable of learning new language nuggets called concepts, directly from end users using live interactive teaching sessions. The proposed setup uses three models to: a) Identify gaps in understanding automatically during live conversational interactions, b) Learn the respective interpretations of such unknown concepts from live interactions with users, and c) Manage a classroom sub-dialogue specifically tailored for interactive teaching sessions. We propose state-of-the-art transformer based neural architectures of models, fine-tuned on top of pre-trained models, and show accuracy improvements on the respective components. We demonstrate that this method is very promising in leading way to build more adaptive and personalized language understanding models.
\end{abstract}

% ===============================================================
\section{Introduction}
\vspace{-4mm}
Humans are adaptive by nature. While it is very natural for humans to ask clarifying questions and immediately correct course whenever there is a misunderstanding, interactive learning using self-supervision has been a holy grail area for Conversational AI. Most Conversational AI systems rely on knowledge-based or machine-learning based (or hybrid) understanding components under the hood to understand the intents of the users’ and their arguments (usually called as slots), such as “{\em set an alarm for 7 am}”, where the intent can be {\em set\_alarm}, and {\em 7 am} can be the time slot. These systems typically do not have the capability to interact with end-users during a live conversation to seek explanations to improve the AI system's understanding of entities, intents or other constructs. When users speak an utterance outside the comprehension of these systems, the AI agent responds with a dead-end response such as ‘Sorry I don’t know that’.

In this paper we describe a teachable AI system towards enabling users to directly teach a conversational AI agent via a live interactive teaching session using natural language explanations. Learning concept definitions using interactive voice is a challenging problem due to a number of reasons, including wide vocabulary of words that users could use to describe definitions, user distractions during teaching sessions, grounding of related entities, fuzziness in the re-use of previously taught concepts across domains, etc. The method described in this paper augments goal-oriented AI agents with an interactive teaching capability, to improve task completion rates by asking questions to fill gaps in the AI agent’s understanding, hence making the agent more personal, conversational and adaptive.

More specifically, the AI agent will be checking the utterances whether it is “Teachable” using a deep learning based parser, which not only classifies the utterance, but also identifies the entity or intent which needs to be learnt from the users. If so, the “Classroom” sub-dialogue kicks in, where the users are asked what they mean by these concepts. This sub-dialogue is driven by a dedicated dialogue manager policy model which incorporates a definition understanding model to interpret users' responses. For example, if the utterance is "{\em set an alarm for my baseball practice}", the system would ask "{\em when is your baseball practice?}".

Despite a few academic papers towards learning directly from users as presented in Section~\ref{sec:rw}, to the best of our knowledge, there is no work that covers all of these novel contributions:
%\begin{itemize}[leftmargin=*]
%\item Ability to use open vocabulary natural language during interactive teaching sessions, 
%\item Ability to guardrail multi-turn conversational dialogue sessions to aid in goal-completion, 
%\item Ability to identify gaps in the agent’s understanding in a wide variety of natural language expressions,
%\item Ability to scale and generalize across multiple domains supported by the AI agent (such as Music, Calendar, Shopping, etc.),
%\item Methodology to re-use the learnt concepts in subsequent conversations with related context, and 
%\item Use the power of transformer based pre-trained language models such as BERT to aid in different aspects of the teaching paradigm.
%\end{itemize}
%The main contributions of this paper are
\begin{itemize}
\itemsep -0.5ex
    \item A dedicated multi-turn domain-agnostic dialogue system specifically tailored for interactive teaching, that augments an existing Conversational AI System to learning explanations from user in real-time.
    \item A multi-task neural Concept Parser that automatically identify gaps in an AI agent’s understanding, using a multi-task model that incorporates semantic role labeling and constituency parser.
    \item A neural Definition Understanding system along with a policy model to aid in conducting robust teaching sessions with the user. 
\end{itemize}
    
The following sections are organized as follows. Section~\ref{sec:rw} describes the related work in the area of interactive learning methods to learn explanations, and predicting gaps in the AI agent’s understanding for a given user input; Section~\ref{sec:approach} describes our architecture and modeling methods, and Section~\ref{sec:exp} describes our experimental evaluations.
% ====================================================================================
\section{Related Work}
\vspace{-2mm}
\label{sec:rw}

The related work on learning via user-in-the-loop for Conversational AI systems can be analyzed in two categories. The higher level interactive learning approaches, and utterance parsing based gap prediction studies.

% \textbf{Interactive Learning}
\subsection{Interactive Learning}
\vspace{-2mm}
Existing work on interactive learning methods can be grouped into 3 categories:
\begin{itemize}
    \item Game learning using limited/no vocabulary: Several methods have been proposed for learning a game (such as Hanoi) using interactive sessions with the user Kirk~{\em et al.}~\cite{kirk-19} Wang~{\em et al.}\cite{wang-17}. Such methods represent the game scene as a symbolic representation of attributes (such as shape, color) which is used to learn the state progression, to learn the game. 
    \item Neural code synthesis methods: Methods like Yin~{\em et al.}~\cite{yin-2020} aim at converting user utterances directly into a coding language like Python. To our knowledge, such syntax-driven code generation methods are still in their early stages, and not yet ready for integration into mainstream conversational AI agents that cover a wide variety of domains. 
    \item Learning by GUI demonstrations: GUI based methods use semantic parsers to learn conditional expressions and their associated actions. One notable study is by Allen~{\em et al.} using a web browser to teach tasks such as buying a book from Amazon~\cite{plow}.
\end{itemize}

\subsection{NLU gap prediction}
\vspace{-2mm}

% https://robinjia.github.io/assets/pdf/icassp2017-concepts.pdf

The problem of predicting segments of an utterance which a base NLU system could not interpret, was tackled by Jia {\em et al.}~\cite{jia-2017}, by using a set of post-processing rules on top of the base NLU's slot-filling model. The rules were built on heuristics from the base NLU model, such as confidence score threshold, out-of-vocabulary flag, and span-expansion based on syntactic dependency tree. This paper also introduced a new dataset for this problem that is publicly available, which is used in our work as one of the datasets for evaluating model performances. In \cite{Kobayashi-2018,kobayashi2019slot} the authors proposed augmenting the training data for the slot-filling model, by injecting noise tokens into the regular slot values of training data, to force the model to learn about the context of the respective slots. This augmentation is done to improve the model robustness in predicting the correct span, on unseen slot values. One variant of such approach uses negative sampling to train the model to identify unknown slot values together with a joint slot tagging and slot classifier~\cite{Hou-2019}. Another line of work focuses on different model structures such as pointer-networks, to better copy unknown slot values from the utterance into final slot value prediction~\cite{xu-2018,Zhao-2018,Yang-2020}. This line of work usually considers the problem under a dialogue state tracking setting, therefore only one slot value is predicted each time. 
% ====================================================================================

\subsection{Definition Understanding}
\vspace{-2mm}
Our task of understanding the definition of concepts from users' explanations, generally falls into the categories of  reading comprehension and slot filling. For reading comprehension, there is a particular thread that formulates question answering as span prediction. Those works typically learn a representation for the question and passage, and then predict the start and end of the answer with attention mechanism between question, passage and answer \cite{chen2017reading,chen2018neural}. The slot filling thread work focuses on sequential labeling of an answer\cite{mesnil2014using}, with most recent works utilizing neural slot-fillers with joint intent classifiers \cite{liu2016attention,goo2018slot,xu2013convolutional,zhang2018joint,chen2019bert}.

% ====================================================================================
\subsection{Dialog Policy for Teaching sessions}
\vspace{-2mm}
While there is relatively rare work for dialog policy models specifically meant for teaching sessions, a dialog policy in general can be learned using a) supervised policy learning, and (b) reinforcement policy learning. Existing supervised learning approaches usually considers it as an intent prediction problem. Recent work usually takes a joint prediction for both intents and slot prediction \cite{liu2016attention,goo2018slot,xu2013convolutional,zhang2018joint,chen2019bert}. Reinforcement learning approaches formulate the policy learning as sequential decision making process, and learns optimal policy from a large number of user interactions \cite{liu2017iterative,takanobu2019guided,zhou2017end,zhao2016towards}.
% ====================================================================================

\section{Methodology}
\vspace{-2mm}
\label{sec:approach}
In this work, we incorporate a dedicated dialogue system called Teachable Dialogue System, which is specifically tailored for conducting interactive teaching sessions with users, for learning definitions of concepts that were previously not understood by the AI system. This dialogue system consists of three components:
%==================
\begin{itemize}[leftmargin=*]
\itemsep -0.5ex
\item Concept parser - that uses pre-trained embeddings and signals from the Conversational AI system to predict gaps in NLU’s understanding for interpreting a given utterance from the user, 
\item Definition understanding model that extracts and interprets explanations from the user and maps the unknown concept phrases to the learnt definitions, and 
\item Dialog policy model that drives teaching conversations with a goal-completion objective of extracting necessary explanations from the user and ground the respective unknown concepts.
\end{itemize}
%==================
\begin{figure}[t]
\centering
\includegraphics[width=0.9\textwidth]{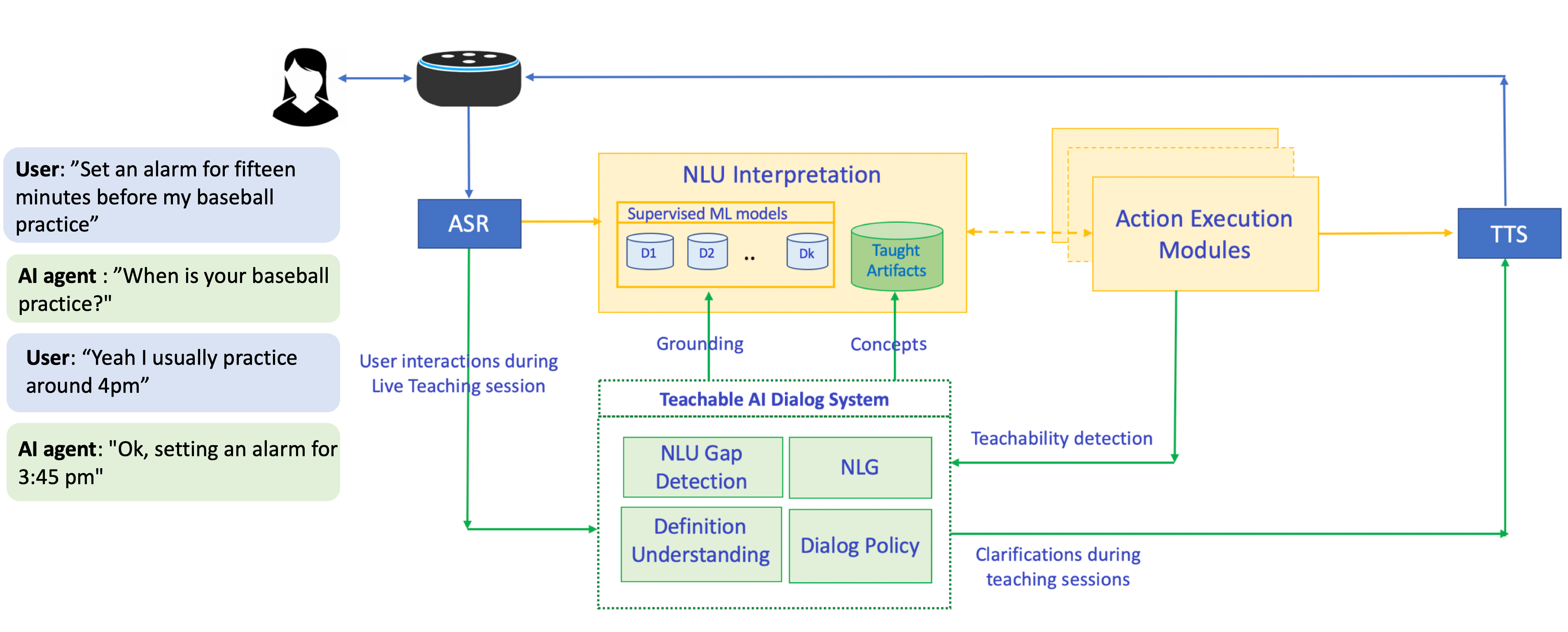}
\caption{Architecture of AI system using a Teachable AI Dailog system}
\label{fig:main}
\vspace{-1.5em}
\end{figure}
%==================
The teachable dialogue system acts as a subsidiary to the multi-domain Conversational AI system, and springs into action when the AI agent is not able to successfully interpret/execute a user’s input utterance. This way, the parent Conversational AI system remains decoupled from the Teachable Dialogue System, staying independent of user-specific interpretation or modeling. Concept parser identifies the sequence of tokens (called concept phrases) that are not understood by the parent NLU system, and helps in initiating a teaching session with the user with the question - "{\em Can you teach me what you mean by <concept phrase>?}" Once a teaching session has been initiated with the user by the dialogue system, the dialog-policy model helps in predicting the right clarification question to ask the user, based on the context of the original utterance and the interactions with the user during the teaching session. The answers from the user are then processed by the definition understanding component that works along with the policy models to conduct multi-turn conversations with the user to learn, clarify and ground the respective concepts in the user’s utterance. When the Teachable Dialogue System deems a teaching session as successful, the taught actions are subsequently executed by the AI system, and the respective definitions and artifacts are then stored for future re-use. Figure~\ref{fig:main} describes the architecture of this Teachable AI system.
\vspace{-2mm}
\subsection{Concept Parser}
\vspace{-2mm}
Identifying gaps in NLU interpretation of an utterance, is a challenging task by itself. This task involves accurately localizing the segments within an utterance that an NLU system is unable to comprehend, and needed to execute the relevant action. In addition to using a slot-tagger objective in our Concept Parser model, we added an auxiliary task of semantic-chunking to make sure the model is sensitive to segments that might be neglected by a token-level slot tagger. For example, the chunking model should be able to segment the sentence: "[show] [me] [navigation] [to] [where we go camping every year]" while the slot tagger may not correctly tag "where we go camping every year" as "Location" concept.

\begin{figure}[tb]
\centering
\includegraphics[width=0.8\textwidth]
{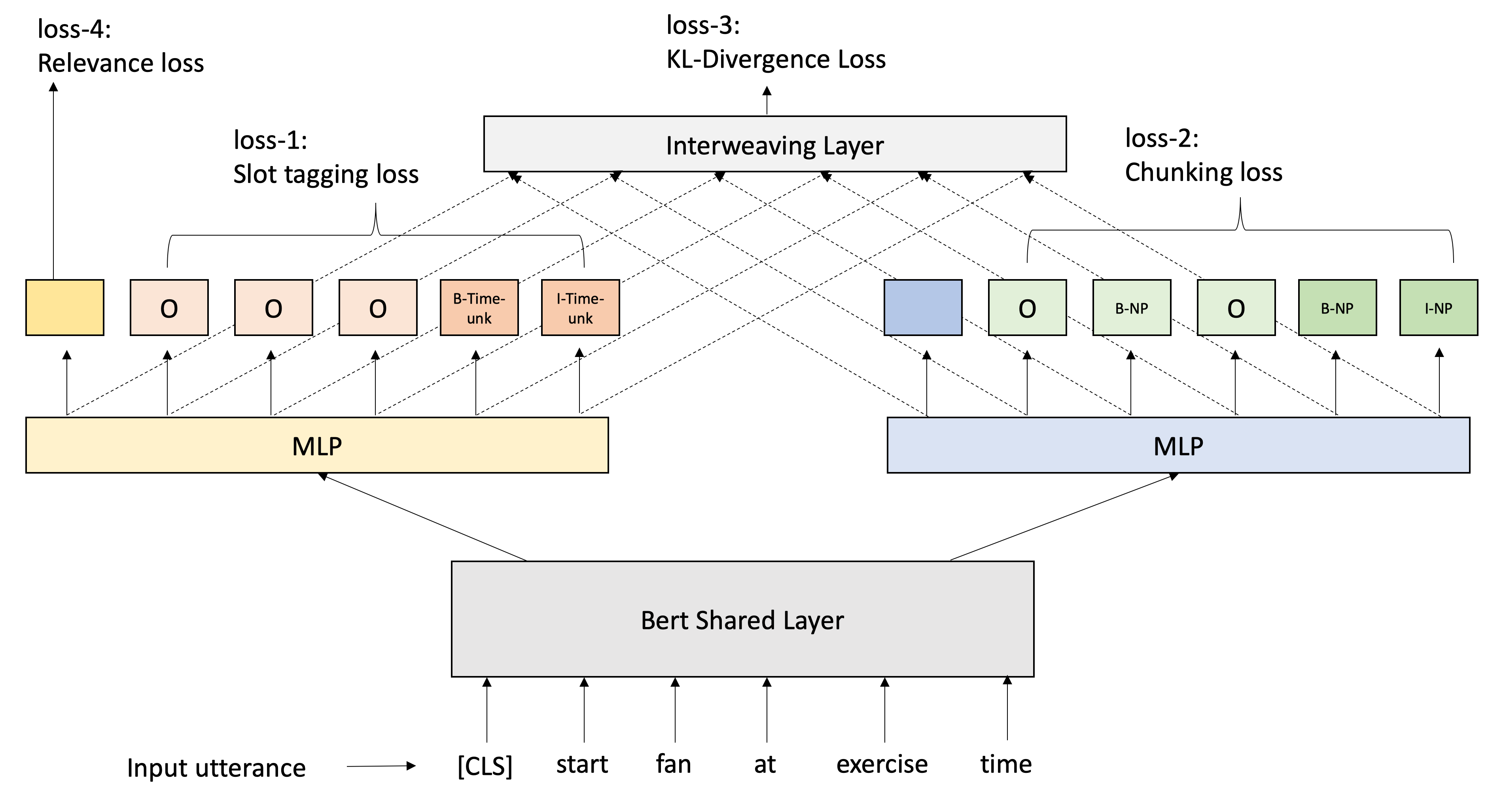}
\caption{
\small
Concept Parser Model Structure: Transformer-based shared layer with Multi-Task Heads
}
\label{fig:cp}
\vspace{-1.0em}
\end{figure}

Additionally, there are several real-world challenges that a Concept Parser needs to address, such as: a) Tentative user behavior - such as "set the lights to, never mind" - where the user intends to cancel the request, rather than mean to set the lights to a user defined value, b) Unsupported actions such as "set the light to fifty degrees" where NLU interpretation is successful, but the Action Execution module cannot execute the respective action - this could be because of an unintentional user error, or due to an error in Automatic Speech Recognition (ASR). Concept Parser needs to treat the above scenarios as 'non teachable' instances, and hence should prevent the initiation of teaching sessions to learn the respective definitions from users. In addition to the task of identifying the concept phrase from a given utterance, Concept Parser also generates 'relevance scores' that indicates how confidence score for the concept phrase to be teachable.

To address the above challenges, we implemented a Concept Parser using a multi-task transformer model which is end-to-end trainable using the following task objectives: a) Slot tagging b) Semantic chunking, and c) Concept Relevance. The model structure is depicted in Figure \ref{fig:cp} , and is composed of the following sub-components:
%==================

\textbf{BERT layer}. It is a multi-layer multi-head  self-attention Bert model \cite{bert}. The input sentence is first tokenized with WordPiece \cite{schuster2012japanese}. Then positional embedding and segment embedding are added to the token embedding as final input embedding $X = {x_1,x_2,...,x_L}$. The final output of bert layer $H = {h_1,h_2,...,h_L}$ is a contextual representation for each input token.

\textbf{Slot tagging head}. Given the bert output $H = {h_1,h_2,...,h_L}$, the slot tagging head feeds the output into a multi-layer perceptron followed by a softmax to predict slot label for each token.
\begin{equation}
z_i = f (W \cdot h_i + b) 
\end{equation}
\begin{equation}
\hat{y}_{ij}(st) = \frac{e^{z_{ij}}}{\sum_{k=1}^{N_1}  e^{z_{ik}} }
\end{equation}
%==================
Where $N_1$ is the number of slot classes and $L$ is the length of the utterance. Then the slot tagging loss $L_{st}$ will be a average cross-entropy loss across all tokens.
\begin{equation}
L_{st} = -  \frac{1}{L \cdot N_1 } \sum_{i=1}^{L } \sum_{j=1}^{N_1} y_{ij}(st) \cdot log (\hat{y}_{ij}(st)) 
\end{equation}
%==================
\textbf{Chunking tagging head}. The chunking head has exact same structure as the slot tagging head. The only difference is the different $W$ and $b$ parameters, as well as the supervised chunking labels $y_i$. The loss $L_{ck}$ can be calculated as follows:
%==================
\begin{equation}
L_{ck} = -  \frac{1}{L \cdot N_2 } \sum_{i=1}^{L } \sum_{j=1}^{N_2} y_{ij}(ck) \cdot log (\hat{y}_{ij}(ck)) 
\end{equation}
%==================
Where $N_2$ is the number of chunking classes and $L$ is the length of the utterance. $y_{ij}(ck)$ is ground-truth chunking labels, and $\hat{y_{ij}(ck)}$ is the predicted chunking labels. We also add an interweaving loss to enforce the "synchronization" between the slot tagging head and chunking head, by imposing a KL-divergence loss on the output of the two heads. 
%==================
\begin{equation}
L_{kl} = - \frac{1}{L \cdot N}\sum_{i=1}^{L} \sum_{j=1}^{N} \hat{z}_{ij}(ck) \cdot log (\frac {\hat{z}_{ij}(ck)}{\hat{z}_{ij}(st)}) 
\end{equation}
%==================
\textbf{Relevance scoring head}. The relevance scoring head takes the bert output embedding of the ${[CLS]}$ token as the input, and feeds it into multi-layer perceptron followed by a binary cross-entropy loss:
\begin{equation}
z_{[CLS]} = f (W \cdot h_{[CLS]} + b) 
\end{equation}
\begin{equation}
\hat{y}_{[CLS],i} = \frac{e^{z_{[CLS],i}}}{\sum_{j=1}^{2} e^{z_{[CLS],j} }}
\end{equation}
%==================
\begin{equation}
L_{rel} = -  \frac{1}{2} \sum_{i=1}^{2} y_i(rel) \cdot log (\hat{y}_{[CLS],i}) 
\end{equation}
%==================
\textbf{Final loss objective}. The final loss objective $L_{CP}$ is a weighted sum of all four losses mentioned above, with $\alpha_1$, $\alpha_2$, $\alpha_3$ and $\alpha_4$ as the weights.
\begin{equation}
L_{CP} = \alpha_1 \cdot L_{st} + \alpha_2 \cdot L_{ck} + \alpha_3 \cdot L_{kl} + (1 -\alpha_1 - \alpha_2 - \alpha_3 ) \cdot L_{rel} 
\end{equation}
%==================
%==================

% \iffalse
% \begin{figure}
% \centering
% \begin{minipage}{0.5\textwidth}
%   \centering
%   \includegraphics[width=0.9\linewidth]{./figures/cp_model_structure.png} \\
%   \captionof{\small(A)}
%   \label{fig:cp}
% \end{minipage}%
% \begin{minipage}{.5\textwidth}
%   \centering
%   \includegraphics[width=0.9\linewidth]{./figures/final_au_model_updated_final.png} \\
%   \captionof{\small(B) }
%   \label{fig:au}
% \end{minipage}
% \caption{
% Architectures of Concept Parser and Definition Understanding Models. Left: (A) Concept Parser: Transformer-based shared layer with Multi-Task Heads. Right: (B) Definition Understanding: A Transformer-based joint intent and span classification model.
% }
% \label{fig:cp_au}
% \end{figure}
% \fi

%==================
\subsection{Definition Understanding}
\vspace{-2mm}
Once Concept Parser has detected a teachable phrase, a teaching session is initiated by requesting the user to provide a suitable definition of the identified concept. There are several real-world challenges involved in accurately extracting the appropriate definition from the user's answer, such as: a) Verbose answers: When posed with an impromptu definition of a concept, users generally tend to use verbose and indirect answers such as ``yeah i mean red color or may be just orange would do'' or ``i meant red not blue''; b) Distracted users: It is quite likely that users do not intend to answer the definition question, but would like to move on with another new request to the AI assistant, e.g. ``Whats the weather outside'', which is not really an answer to the question; c) Incomplete answers: Users might have provided an answer but its not yet sufficient to fully ground the respective definition and needs more clarification questions; d) Complicated answers: Users could provide a definition that sounds logical to another human, but the respective definition could still be not be understood by the parent AI system and hence cannot be grounded to the equivalent actions; e) Contextual references: such as ``make it brighter''.

\begin{figure}[t]
\centering
\includegraphics[width=0.75\textwidth]
{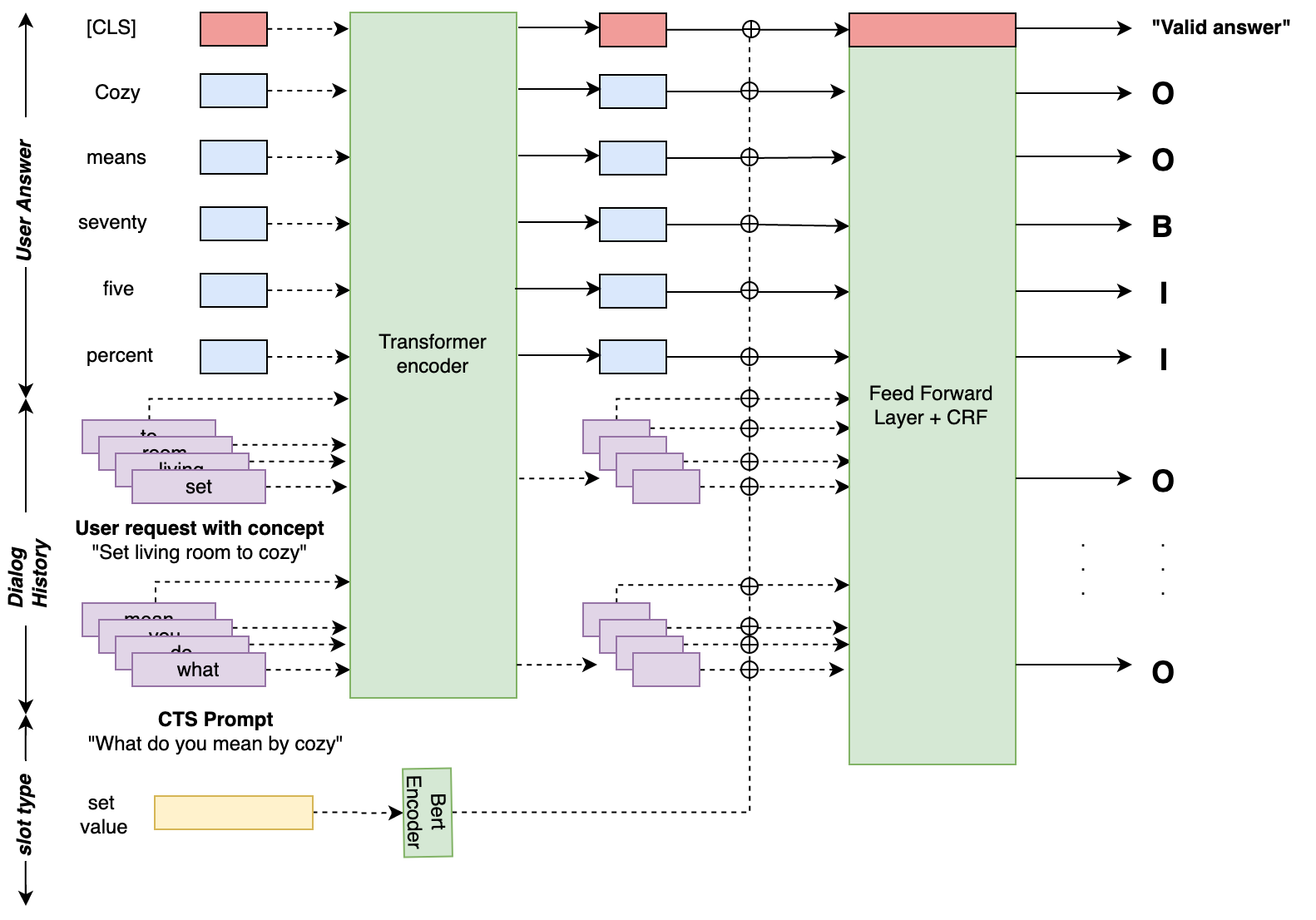}
\vspace{-1.0em}
\caption{Definition Understanding model}
\label{fig:au}
\vspace{-1.5em}
\end{figure}
% \begin{figure}
%     \resizebox{.9\linewidth}{!}{./figures/final_au_model_updated_final.png}
% \end{figure}

% We developed a Definition Understanding model that can generically address the above scenarios, that uses the entire dialogue history from the conversation, to both extract the concept definition and also to determine if the user has provided a valid answer within the teaching session, or made a new request to the AI system by ignoring the clarification questions.
We developed a Definition Understanding model that can generically address the above scenarios. The model uses a Transformer-based joint intent and span detection framework that takes in the dialogue history ($\bm{z}$ = $z_1$, $z_2$, $\dots$, $z_{H}$), user answer ($\bm{x} = x_1, x_2, \dots, x_{A}$) and the slot type ($\bm{m} = m_1, m_2, \dots, m_{S}$), illustrated in Figure~\ref{fig:au}. We first concatenate user answer and dialogue history and pass the concatenated vector through a Transform-based encoder (e.g. BERT~\cite{bert}) and extract the encoder's last layer hidden states, denoted as $\bm{H} = (\bm{h}_{[CLS]}, \bm{h}_{x, 1}, \dots, \bm{h}_{x, A}, \bm{h}_{[SEP]}, \bm{h}_{z, 1}, \bm{h}_{z, 2}, \dots, \bm{h}_{z, H}, \bm{h}_{[SEP]})$. Separately, we obtain a slot type embedding, $\bm{e}_s$ with some pre-trained model (e.g. BERT). The contextual representation of each token is then fused with the slot type embedding

%----
\begin{equation}
    \bm{H} \oplus \bm{e}_s = (\bm{h}_{[CLS]} \oplus \bm{e}_s, \dots, \bm{h}_{x, i} \oplus \bm{e}_s, \dots \bm{h}_{z, j} \oplus \bm{e}_s, \dots, \bm{h}_{[SEP]} \oplus \bm{e}_s)
\end{equation}
%----
and further gets passed through a set of post-Transformer encoder layers, e.g. feed-forward and CRF layers, to produce the final representation of each input token, $\bm{H}^O = (\bm{h}_{[CLS]}^O, \dots, \bm{h}_{x, i}^O, \dots \bm{h}_{z, j}^O, \dots, \bm{h}_{[SEP]}^O)$. Intent and span classification are performed with separate output layers and described below in details.
%----

\textbf{Intent classification}. We adopt a single fully connected layer followed by a softmax layer on top of the final representation of ${[CLS]}$ token to perform the intent classification.

\begin{equation}
    \hat{y}_i^{intent} = {Softmax}(f(\bm{W}^{intent} \cdot \bm{h}_{[CLS]}^O + \bm{b}^{intent}))
\end{equation}
We use cross-entropy to calculate the intent classification loss for $C$ intents.

\begin{equation}
    L_{intent} = - \frac{1}{C} \sum_{i=1}^C y_i^{intent} \cdot \log({\hat{y}_i^{intent}})
\end{equation}
%----
\textbf{Span classification}. We use the same layer structure to perform span classification as intent classification except that the fully connected layer is 3-way (i.e. ``B'', ``I'', ``O'').

\begin{equation}
    \hat{y}_{i,k}^{span} = {Softmax}(f(\bm{W}_i^{intent} \cdot \bm{h}_{x, i}^O + \bm{b}_i^{span}))
\end{equation}
%----
The span classification loss is also calculated using cross-entropy:

\begin{equation}
    L_{span} = - \frac{1}{3A} \sum_{i=1}^A \sum_{k=1}^3 y_{i,k}^{span} \cdot \log({\hat{y}_{i,k}^{span}}) - \frac{1}{3H} \sum_{j=1}^H \sum_{k=1}^3 y_{j,k}^{span} \cdot \log({\hat{y}_{j,k}^{span}})
\end{equation}
%----
\textbf{Joint loss}. The final loss objective $L_{DU}$ is a weighted sum of intent and span classification losses. The hyperparameter of relative intent loss proportion $\alpha_{intent}$ is furthered tuned on a validation dataset.

\begin{equation}
    L_{DU} = \alpha_{intent} \cdot L_{intent} + (1 - \alpha_{intent}) \cdot L_{span}
\end{equation}

\subsection{Dialogue Policy}
\vspace{-2mm}
For the Teachable Dialogue system described in this paper, a Dialogue Policy predicts the next action in a multi-turn teaching dialogue session with a goal-completion target of extracting all the necessary definitions from the user during the teaching. A successful completion of the teaching session is one where the concept phrase in the first-turn utterance has been grounded to an equivalent action that is executable by the parent AI system. 
The Dialogue Policy uses a Transformer-based contextual model to predict the next action during a teaching session. The action-space for our Dialogue Policy model includes a) Ask or repeat a clarification question, b) Guardrail the conversations to channelize users back into the teaching conversations, c) Identify Out-Of-Domain (OOD) turns during a teaching session, d) Ground the extracted definitions with the parent NLU system, and e) deem a teaching session as successful or unsuccessful, or decide to end a teaching session. The dialogue policy model takes in utterances from contextual dialogue history, predicted definition spans and confidence scores from Definition Understanding module, and the set of recognized/resolved slots from the parent NLU model and passes the input through a Transformer-based encoder to get the last layer output of ${[CLS]}$ token as a contextual representation of the inputs, $\bm{h}_{[CLS]}$. Then we apply a single feed-forward layer and a softmax layer to obtain a probability distribution over all the possible actions.
\begin{equation}
    \hat{y}^{action} = {Softmax}(f(\bm{W}^{action} \cdot \bm{h}_{[CLS]} + \bm{b}^{action}))
\end{equation}

\section{Experiments and Results}
\label{sec:exp}
\vspace{-3mm}
% ========================================================================
% \subsection{Datasets}
% \label{sec:datasets}
\textbf{Datasets:} We evaluate our results on two datasets: 1) The dataset published by \cite{jia-2017} which consists of both personal concepts and generic slot values for five slot types namely date, time, location, people and restaurant-name. While the dataset in \cite{jia-2017} pertains to the problem that we have attempted to solve in this paper, we find that it does not include challenging real-world scenarios such as the ones described in Section~\ref{sec:approach} which are addressed by our models. We hence used a second internal dataset collected from crowd-sourcing which addressed the challenges of realistic multi-turn teaching sessions. Crowd workers provided both cooperative and non-cooperative inputs for two tasks: 
\begin{itemize}[leftmargin=*]
\itemsep -0.5ex
    \item Generate first-turn utterances containing both personal and non-personal concept phrases for performing actions supported by the AI assistant spanning multiple domains. Each utterance was annotated with the ground-truth concept phrase. For the "not teachable" class, we synthesize a wide variety of cases, including utterances without any concept, out-of-domain utterance, ill-grammar and incomplete utterances.
    \item Answers to clarification questions within teaching sessions. Answers were annotated with the concept definition phrases along with additional information like direct-answer, new-request, etc. 
\end{itemize}
%===========
\textbf{Concept parser experiment setting}: For concept parser model, a pre-trained BERT-base model is used as the base layer, and it is fine-tuned together with all the head layers during training \cite{bert}. We used a 300 dimensional hidden-layer in MLP, a RELU activation function \cite{nair2010rectified}, and a learning rate of 1e-5. For the public dataset, the model is simultaneously trained on $L_{st}$, $L_{ck}$ and $L_{kl}$ losses without relevance scoring for 20 epochs, where weights are set to 0.5/0.5/2.0 empirically. For our internal dataset, the model is first trained on the $L_{ck}$ loss for 2 epochs, and then fine-tuned on $L_{st}$ and $L_{rel}$ losses with 1.0/1.0 weights for 2 epochs. For the auxiliary chunking label, we use a fusion of semantic role labeling model(SRL) and constituency parsing model (CPM) models from AllenNLP \footnote{https://allennlp.org/}.

\textbf{Definition understanding experiment setting}: We experimented with a few pre-trained transformer-based models, namely BERT~\cite{bert}, RoBERTa~\cite{roberta}, ALBERT~\cite{albert} as the base encoder. We used a single feed-forward layer with hidden state set to 100 and RELU \cite{nair2010rectified} activation function followed by a CRF layer for post-Transformer encoder. The initial learning rate is set to $1 \times 10^{-4}$ and reduces to half at loss plateau, i.e. no loss reduction within last consecutive 10 epochs. We used Adam optimizer~\cite{adam} with weight decay enabled.

% ========================================================================
\subsection{Performance evaluations}
\vspace{-2mm}
\subsubsection{Concept Parser results}
\vspace{-2mm}

\textbf{Public Datasets}: Table~\ref{tab1} shows the performance of our concept parser on personalized test set from \cite{jia-2017} using phrase-level metrics on slots with unknown concepts. From Table~\ref{tab1} we observe the following: a) When there is no personalized training data available (Zero-shot), there is an absolute improvement in F1-score of 8.6\% for single-task model, and 15\% using Multi-task model. This supports our idea that an end-to-end model with built-in objectives for both slot tagging and semantic-chunking can be more robust in this zero-shot unknown concept prediction setting. b) When personalized concept phrases are added to the training data, the model performance jumps to 70\% and 76\% for single-task and multi-task models  respectively. Some source of errors still remains due to annotation inconsistencies ("after lunch" or "lunch" as concept).

%=======
\begin{table}
    \begin{minipage}[t]{0.5\linewidth}
    %   \centering
        \begin{tabular}[t]{  lcccc }
        \cmidrule[\heavyrulewidth]{1-4}
        \textbf{Model} & \textbf{Precision} & \textbf{Recall} & \textbf{F1}\\[0.2em]
        \cmidrule[\heavyrulewidth]{1-4}
        \textbf{Zero-shot} \\
            Jia {\em et al.}~\cite{jia-2017}  & 41.1\% & 40.0\% & 40.3\%\\ 
            \cmidrule{2-5}
            Single-Task BERT           & 44.0\% & 55.1\% & 48.9\%\\ 
            \cmidrule{2-5}
            Multi-Task BERT            & \textbf{50.5}\% & \textbf{61.3}\% & \textbf{55.3}\%\\ 
        \cmidrule[\heavyrulewidth]{1-4}
        \textbf{Supervised} \\
        Jia {\em et al.}~\cite{jia-2017}         & N/A & N/A & N/A\\[0.2em]
            \cmidrule{2-5}
            Single-Task BERT           & 62.2\% & 80.0\% & 70.0\%\\[0.2em]
            \cmidrule{2-5}
            Multi-Task BERT            & \textbf{69.1}\% & \textbf{86.0}\% & \textbf{76.6}\%\\[0.2em]
        \cmidrule[\heavyrulewidth]{1-4}
        \end{tabular}
      \caption{Concept Parser performance on unknown concepts on Personalized Test.}
      \label{tab1}
    \end{minipage}%
    \hspace{3em}
    \begin{minipage}[t]{0.45\linewidth}
    %   \centering
        \begin{tabular}[t]{  p{1.8cm}p{1cm}p{1cm}p{1cm} } 
        \cmidrule[\heavyrulewidth]{1-4}
        \textbf{\space\space\space Model} & \textbf{Precision} & \textbf{Recall}& \textbf{F1}\\[0.2em]
        \cmidrule{1-4}
        \cmidrule{1-4}
        Single-Task BERT  & 90.86 & 91.72 & 91.29 \\[0.2em]
        \cmidrule{1-4}
        Multi-Task BERT   & 93.24 & 84.86 & 88.86 \\[0.2em]
        \cmidrule[\heavyrulewidth]{1-4}
        \end{tabular}
    %   \centering
        \caption{Concept Parser performance on internal dataset using models with additional Relevance head}
        \label{tab2}
    \end{minipage} 
\end{table}

\textbf{Internal Datasets}: Table~\ref{tab2} reports the performance of our concept parser on the internal dataset. From the table we can observe the following. Our single-task and multi-task models, both trained on personalized data, achieve competitive performance on the synthesized evaluation data. When personalized data are available in training data, single-task model performs already decently (91.29\% and 88.86\%); yet multi-task model could still achieve higher precision. We would like to highlight here that our internal evaluation dataset is also more challenging than the public dataset, including a wide variety of "not teachable" examples including out-of-domain requests, regular requests without concept and ill-grammar and incomplete utterances.

% ========================================================================
\vspace{-0.75em}
\subsubsection{Definition understanding results}
\vspace{-0.5em}
%=======
\textbf{Public Datasets}: Table~\ref{tab3} shows the performance of our definition understanding model on personalized test set from \cite{jia-2017} with three Transformer-based encoders. From the results, we observe that model with RoBERTa~\cite{roberta} encoder achieves the best performance in terms of precision, recall and F1-score. BERT~\cite{bert} and ALBERT~\cite{albert} have similar performance and both are lightly behind RoBERTa~\cite{roberta}. We also did an error analysis and found that nearly half of the span prediction errors are due to a mismatch on preposition (e.g. "on Sept 15" vs "Sept 15"), which is not unexpected since span annotations in \cite{jia-2017} are not consistent on the inclusion of prepositions.

\textbf{Internal Datasets}: The performance of our concept parser on the internal dataset is reported in Table~\ref{tab4}. Again, we observe that model with RoBERTa~\cite{roberta} encoder slightly outperforms model with BERT~\cite{bert} and ALBERT~\cite{albert}. 
%We also experimented various post-Transformer layer combinations and %observed a noticeable performance drop by excluding CRF layer. 
%We also found that having an LSTM layer doesn't affect the %performance much since all the three Transformer encoders have %efficiently captured contextual information within the dialog.
%=======
\begin{table}
    \begin{minipage}[t]{0.5\linewidth}
      \centering
        \begin{tabular}{lccc}
        \cmidrule[\heavyrulewidth]{1-4}
        \textbf{Model} & \textbf{Precision} & \textbf{Recall}& \textbf{F1-score}\\[0.2em]
        \cmidrule{1-4}
        RoBERTa             & \textbf{91.80} & \textbf{90.89} & \textbf{91.34}\\[0.2em]
        \cmidrule{1-4}
        BERT                & 90.71 & 90.01 & 90.36 \\[0.2em]
        \cmidrule{1-4}
        ALBERT              & 89.93 & 90.53 & 90.23\\[0.2em]
        \cmidrule[\heavyrulewidth]{1-4}
        \end{tabular}
      \caption{Definition Understanding results on Jia {\em et al.} \cite{jia-2017} dataset}
      \label{tab3}
    \end{minipage}%
    \hspace{2em}
    \begin{minipage}[t]{0.5\linewidth}
      \centering
        \begin{tabular}{lccc}
        \cmidrule[\heavyrulewidth]{1-4}
        \textbf{Model} & \textbf{Precision} & \textbf{Recall}& \textbf{F1-score}\\[0.2em]
        \cmidrule{1-4}
        RoBERTa             & \textbf{95.92} & \textbf{96.81} & \textbf{96.36}\\[0.2em]
        \cmidrule{1-4}
        BERT                & 95.57 & 96.03 & 95.80\\[0.2em]
        \cmidrule{1-4}
        ALBERT              & 95.41 & 95.08 & 95.24\\[0.2em]
        \cmidrule[\heavyrulewidth]{1-4}
        \end{tabular}
      \centering
        \caption{Definition Understanding results on internal dataset}
        \vspace{-1.2em}
        \label{tab4}
    \end{minipage}
\end{table}

%====
\subsubsection{Policy Model results}
\vspace{-2mm}
Table~\ref{tab5} shows the performance of our Transformer-based policy model along with heuristics based on NLU grounding results, on an internal dataset with annotated ground-truth actions. We observe that model with RoBERTa~\cite{roberta} encoder achieves the best performance in terms of precision, recall and F1-score. ALBERT~\cite{albert} showed comparable performance to BERT~\cite{bert} but ran the fastest.
%====
% ========================================================================
\begin{table}[h]
\centering
\begin{tabular}{lccccccccc}
\cmidrule[\heavyrulewidth]{1-5}
\textbf{\space\space\space Model} & \textbf{Precision} & \textbf{Recall}& \textbf{F1}& \textbf{mins/epoch}\\[0.2em]
\cmidrule{1-5}
RoBERTa & \textbf{98.78} & \textbf{97.71} & \textbf{98.24} & 6.12\\[0.2em]
\cmidrule{1-5}
BERT  & 98.73 & 96.26 & 97.48 & 5.11\\[0.2em]
\cmidrule{1-5}
ALBERT   & 97.98 & 96.12 & 97.04 & 4.56\\[0.2em]
\cmidrule[\heavyrulewidth]{1-5}
\end{tabular}
\caption{Policy Model performance evaluation results on internal dataset}
\vspace{-1.2em}
\label{tab5}
\end{table}

% ========================================================================
% ========================================================================
\vspace{-3mm}
\section{Conclusion}
\vspace{-3mm}
We have presented a teachable dialogue system that uses neural models for gap identification, definition understanding and dialogue policy prediction to conduct interactive teaching sessions with the user to learn and re-use definitions of concepts that are unknown to a conversational AI system. This Teachable Dialogue system helps in automatically improving the understanding capabilities of the AI system to hold more natural conversations with the end-users, and progressively improve the agents’ understanding of users' parlance to enable more natural ways of interaction with a conversational AI system. We believe this is an effort towards building truly interactive learning systems, and plan to extend the scope to higher level capabilities such as intent teaching.

%\begin{ack}
%We would like to thank ...
%\end{ack}

\bibliographystyle{plain}
\bibliography{references.bib}

\end{document}